\documentclass[letterpaper]{article} 
\usepackage{aaai2027}  
\nocopyright
\usepackage[hyphens]{url}  
\usepackage{graphicx} 
\usepackage{natbib}  
\usepackage{caption} 
\usepackage{amsmath} 
\usepackage{amssymb} 
\usepackage{algorithm}
\usepackage{algorithmic}
\usepackage{booktabs,multirow,graphicx,array}

\usepackage{newfloat}
\usepackage{listings}
\DeclareCaptionStyle{ruled}{labelfont=normalfont,labelsep=colon,strut=off} 
\floatstyle{ruled}
\newfloat{listing}{tb}{lst}{}
\floatname{listing}{Listing}

\usepackage{booktabs}
\usepackage{array}   
\usepackage{multirow}

\title{AsyTO: Asymmetric Temporal Operator for Parameter-Efficient \\ Multivariate Time Series Forecasting}
\author{
    Xiachong Lin\textsuperscript{\rm 1},
    Du Yin\textsuperscript{\rm 1},
    Hao Xue\textsuperscript{\rm 2},
    Wen Hu\textsuperscript{\rm 1}, 
    Imran Razzak\textsuperscript{\rm 3}, 
    Arian Prabowo\textsuperscript{\rm 1}, \\
    Matthew Amos\textsuperscript{\rm 4}, 
    Flora D. Salim\textsuperscript{\rm 1}
}
\affiliations{
    \textsuperscript{\rm 1}University of New South Wales \\
    \textsuperscript{\rm 2} Hong Kong University of Science and Technology (Guangzhou) \\
    \textsuperscript{\rm 3} Mohamed bin Zayed University of Artificial Intelligence \\
    \textsuperscript{\rm 4} CSIRO Energy Centre \\
}

\begin{document}

\maketitle

\begin{abstract}
  Multivariate time-series forecasting faces a structural dilemma: sharing one temporal predictor across variables is parameter-efficient but forces heterogeneous variables through an identical history-to-future map, whereas learning an independent predictor per variable restores flexibility at a cost that grows with the product of variable count, context length, and horizon.
  We argue that this dilemma dissolves once the object being compressed is the forecasting \emph{operator} rather than the observed series. Auditing per-variable linear history-to-future maps across standard benchmarks, we find that a phase-locked seasonal component paired with a compact residual operator
  outperforms a dense phase-blind reference in most audited settings. The residual transport is also \emph{directional}: lag-invariant alternatives consistently underperform asymmetric history-to-future maps. Guided by this structure, we propose AsyTO, an \emph{Asymmetric Temporal Operator} that factorizes the tensor of per-variable operators into shared but distinct history-reading and future-writing temporal modes with per-variable
  mode-wise gains, complemented by a low-rank periodic prototype and a
  cycle-separable factorization of the temporal modes. Each forecast reads only its own variable's history, so parameters and compute grow linearly in the number of variables. Across eleven benchmarks and multiple forecast horizons, AsyTO attains the best lightweight error in $30$ of $44$ dataset-horizon settings, locating at the accuracy-compute Pareto frontier.
\end{abstract}


\section{Introduction}
\label{sec:intro}
Long-horizon multivariate time series forecasting has recently been reshaped by lightweight models. Linear and frequency-domain predictors with modest parameter counts~\cite{zeng2023transformers,xu2024fits,lin2024sparsetsf} match or outperform Transformer-based forecasters~\cite{zhou2021informer,wu2021autoformer,nie2022time} on standard benchmarks. The central question has therefore shifted from whether a forecaster can be small to how its limited parameter budget should be allocated. Compact forecasters spend this budget along both variable and temporal dimensions. Sharing one map across variables is efficient but imposes a common temporal response, whereas learning a dense map for each variable preserves heterogeneity but scales poorly with the panel size. Temporal compression reduces this cost, yet common constructions describe both ends of the forecasting map with the same object, such as basis, kernel, or pattern set. Their efficiency is therefore obtained by tying how historical evidence is read to how the future trajectory is written.
We argue that these roles should remain distinct. Reading the history is an evidence-extraction problem that may concentrate on a few informative lags, whereas writing the forecast requires coordinating the entire horizon. We call this functional distinction \emph{history--future asymmetry}. Preserving it ordinarily requires separate temporal parameters, creating a tension between role specialization and model size. The key question is whether asymmetry can be retained without surrendering compactness.

Our model resolves this tension through periodic structure. A low-rank periodic prototype first estimates the recurring component and produces a residual series. The Asymmetric Temporal Operator (AsyTO) acts exclusively on this residual, using separate history-reading and future-writing factors shared across variables. Variable-specific gains are generated from coordinates learned by the periodic prototype, while cycle-separable factorization compresses both temporal factors. The final forecast combines the periodic component with the predicted residual, and each variable is predicted solely from its own history. Main contributions of this work are organized as below:
\begin{itemize}
    \item We formulate parameter-efficient multivariate time series forecasting as structured compression of the pervariable history-to-future operator tensor, and formalize history–future asymmetry as a distinct budget-aware inductive bias.
    \item We propose \textbf{AsyTO}, which factorizes each variable-specific operator and combines it with a low-rank, phase-aligned periodic prototype and cycle-separable temporal factors. Prototype-conditioned gains encode heterogeneous temporal responses in a shared low-dimensional space.
    \item Extensive structural controls and benchmark evaluations validate the proposed design across several real-world time-series datasets. Our \textbf{AsyTO} improves forecasting accuracy across most datasets while requiring substantially fewer parameters. The proposed modules further improve several popular lightweight backbones in most cases, respectively.
\end{itemize}

\section{Related Works}
\label{sec:related_work}

\paragraph{Compact Forecasting and Temporal Compression.}
Lightweight temporal models range from linear predictors such as DLinear~\cite{zeng2023transformers}, MixLinear~\cite{mamixlinear}, and LightTS~\cite{zhang2022lightts} to sparse or spectral designs including FITS~\cite{xu2024fits}, SparseTSF~\cite{lin2024sparsetsf}, FreTS~\cite{yi2023frequency}, and FilterNet~\cite{yi2024filternet}. An orthogonal distinction concerns variable interaction: iTransformer~\cite{liu2024itransformer} and query-based models~\cite{lin2025temporal} mix information across the panel, whereas PatchTST~\cite{nie2022time} enforces channel independence; this trade-off is studied in~\cite{han2024channelstrategy}. 
Existing approaches also compress the history-to-forecast map through fixed-basis spectral interpolation~\cite{xu2024fits}, relative-lag convolutions~\cite{liu2022scinet}, or learned temporal modes~\cite{ni2023basisformer}.
AsyTO follows channel-independent data flow while sharing its temporal factors across variables, but differs from these compression schemes by assigning separate learned factors to history reading and forecast generation under the same parameter budget.

\paragraph{Explicit Periodic Modeling.}
CycleNet~\cite{lin2024cyclenet} removes a learned cycle before forecasting the residual, while SparseTSF~\cite{lin2024sparsetsf}, PhaseFormer~\cite{niu2025phaseformer}, and FreqCycle~\cite{zhang2026freqcycle} exploit periodicity through downsampling, phase tokenization, and cycle-aligned spectra. AsyTO shares CycleNet's decomposition step and makes no novelty claim for periodic removal. Its distinction is to reuse the periodic prototype to parameterize residual forecasting: prototype coordinates generate variable responses, while cycle--phase structure compresses the asymmetric temporal factors.
\section{Methodology}
\label{sec:method}

\begin{figure*}[t]
    \centering
    \includegraphics[width=\textwidth]{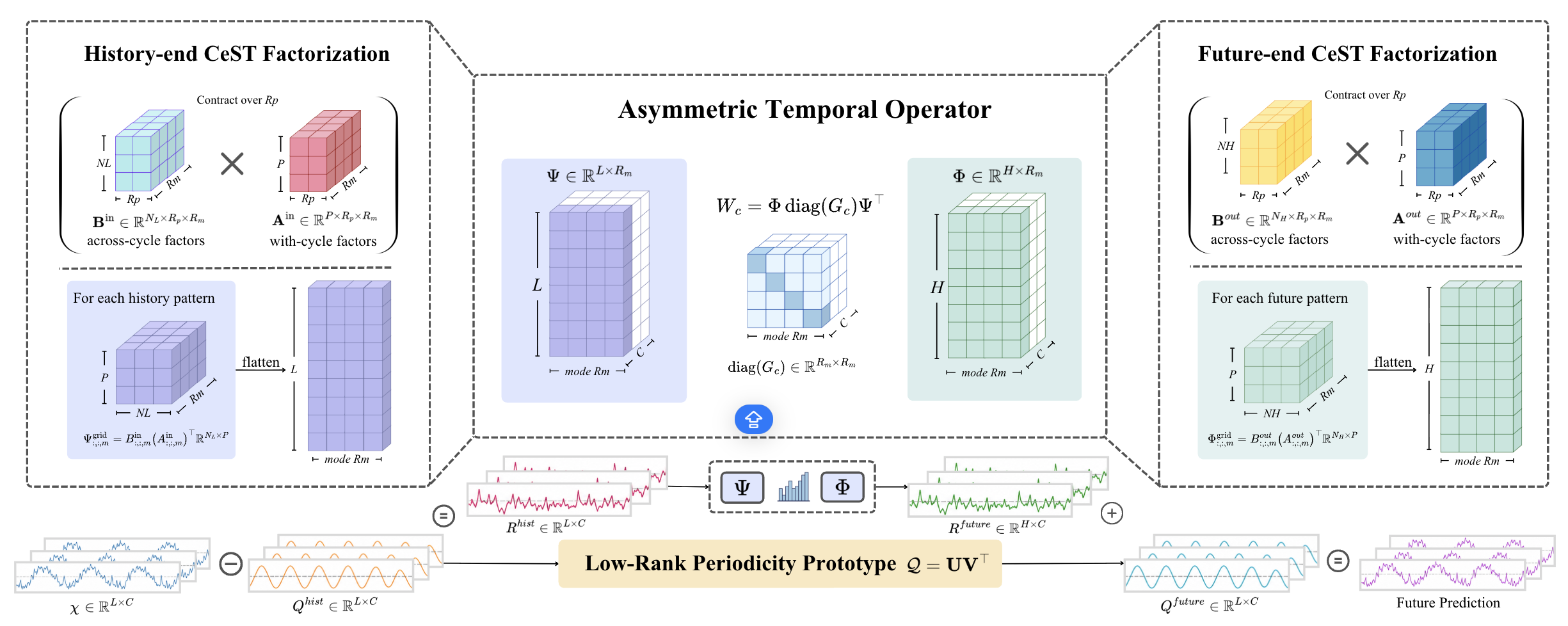}
    \caption{AsyTO. A low-rank periodic prototype $\mathcal{Q}=\mathbf{U}
    \mathbf{V}^{\top}$ is removed from the history and restored on the
    future axis by phase lookup; the periodicity-exclusive residual is
    transported by an asymmetric operator whose history and future factors
    $(\boldsymbol{\Psi},\boldsymbol{\Phi})$ are separate and shared across
    variables, with a per-variable modal response $\mathbf{G}$. Both temporal
    factors are further compressed into a cycle-index and a within-cycle phase factor (CeST), and $\mathbf{G}$ reuses the prototype's variable
    coordinates, so each additional variable costs only $R_p$ parameters.}
    \label{fig:pipeline}
\end{figure*}
\subsection{Problem Formulation}
\label{sec:problem_formulation}
Denote a batch of historical time-series observations $\mathcal{X}\in\mathbb{R}^{B\times L\times C}$, the objective is to predict the subsequent $H$ observations $\mathcal{Y}\in\mathbb{R}^{B\times H\times C}$, where $B$, $L$, and $H$ denote the batch size, context length, and forecast horizon.
For sample $b$ and variable $c$, $\mathcal{X}_{b,:,c}\in\mathbb{R}^{L}$ and $\mathcal{Y}_{b,:,c}\in\mathbb{R}^{H}$ are its historical context and
future trajectory. 
The forecasting problem for the given variable is defined as 
\begin{align}
  \widehat{\mathcal{Y}}_{b,:,c} = f_{\theta}\!\left(\mathcal{X}_{b,:,c};c\right), \qquad c\in\{1,\ldots,C\}.
\label{eq:problem}
\end{align}
where $f_\theta\in\mathbb{R}^{L\times H}$ is the forecaster operator bridging the information transport from history to future.

\subsection{Low-Rank Periodic Prototype}
\label{sec:method_periodic_prototype}
Let $P$ be the dataset period and $s_b\in\{0,\ldots,P-1\}$ the phase of the
first historical observation of sample $b$, so position $\tau$ carries phase
$\pi_b(\tau)=(s_b+\tau-1)\bmod P$. Instead of a dense phase--variable table in
$\mathbb{R}^{P\times C}$, we learn a low-rank prototype
\begin{align}
\mathcal{Q}=\mathbf{U}\mathbf{V}^{\top},\quad
\mathbf{U}\in\mathbb{R}^{P\times R_p},\;
\mathbf{V}\in\mathbb{R}^{C\times R_p},
\label{eq:low_rank_cycle}
\end{align}
where $\mathbf{U}$ holds within-cycle patterns shared by all variables and
$\mathbf{V}_{c,:}$ places variable $c$ in that subspace, reducing the cost
from $PC$ to $R_p(P{+}C)$. The assumption is low rank of the periodic
template, not of the observations themselves.

History and future components are read off the same prototype by phase,
$\mathcal{Q}^{\mathrm{hist}}_{b,t,c}=\mathcal{Q}_{\pi_b(t),c}$ and
$\mathcal{Q}^{\mathrm{future}}_{b,h,c}=\mathcal{Q}_{\pi_b(L+h),c}$. Subtracting
the first leaves the periodicity-exclusive residual
\begin{align}
\mathcal{R}^{\mathrm{hist}}_{b}
=\widetilde{\mathcal{X}}_{b}-\mathcal{Q}^{\mathrm{hist}}_{b}
\in\mathbb{R}^{L\times C},
\label{eq:periodicity_exclusive_residual}
\end{align}
which the operator transports, while $\mathcal{Q}^{\mathrm{future}}$ supplies the phase-aligned future component and will be added back eventually. We initialize $\mathbf{U}$ orthogonally and
$\mathbf{V}$ to zero, so training starts from a neutral template while
$\mathbf{V}$ still receives gradient.

\subsection{Asymmetric Temporal Operator}
\label{sec:method_asy_operator}
The residual $\mathcal{R}^{\mathrm{hist}}$ must now be transported to the
future axis. Giving every variable its own dense map means an operator
$\mathcal{W}\in\mathbb{R}^{C\times H\times L}$ with $CHL$ parameters; sharing
one map across variables is far smaller but forces identical temporal
responses. We interpolate between the two with a CP factorization into three
semantically distinct factors: history-side factors
$\boldsymbol{\Psi}\in\mathbb{R}^{L\times R_m}$, future-side factors
$\boldsymbol{\Phi}\in\mathbb{R}^{H\times R_m}$, and a variable--mode response
$\mathbf{G}\in\mathbb{R}^{C\times R_m}$ whose $c$-th row is $G_c$. For
variable $c$,
\begin{align}
\mathcal{W}_c=\boldsymbol{\Phi}\operatorname{diag}(G_c)\boldsymbol{\Psi}^{\top}
=\sum_{m=1}^{R_m}G_{c,m}\phi_m\psi_m^{\top},
\label{eq:asymmetric_operator}
\end{align}
and $\mathcal{R}^{\mathrm{future}}_{b,:,c}=\mathcal{W}_c\,
\mathcal{R}^{\mathrm{hist}}_{b,:,c}$. Here $\psi_m=\boldsymbol{\Psi}_{:,m}$
reads a pattern out of the historical residual and $\phi_m=
\boldsymbol{\Phi}_{:,m}$ writes the corresponding future trajectory; both are
shared across variables, and $G_{c,m}$ sets how strongly variable $c$
responds to mode $m$.

The operator never has to be materialized. Contracting in mode space,
\begin{align}
Z_{b,m,c}=\sum_{t=1}^{L}\Psi_{t,m}\mathcal{R}^{\mathrm{hist}}_{b,t,c},
\qquad
\mathcal{R}^{\mathrm{future}}_{b,h,c}=\sum_{m=1}^{R_m}\Phi_{h,m}G_{c,m}Z_{b,m,c},
\label{eq:asymmetric_contraction}
\end{align}
costs $R_m(L+H+C)$ parameters and $\mathcal{O}(BC(L+H)R_m)$ arithmetic. It
also makes the target-only property explicit: the forecast for variable $c$
reads $\mathcal{R}^{\mathrm{hist}}_{b,:,c}$ and never another variable's
observations.

Crucially, $\boldsymbol{\Psi}$ and $\boldsymbol{\Phi}$ are parameterized
separately. Even when $L=H$, reading evidence from a past position and
generating a future one are different operations, and tying $\Psi=\Phi$ would
force a symmetric temporal map. The ordered pair $(\Psi,\Phi)$ is what makes
the transport directional, and this asymmetry is the property the operator
cannot give up (Figure~\ref{fig:operator_asy}).

\subsection{Cycle-Separable Temporal Factorization}
\label{sec:method_cest}
Equation~\eqref{eq:asymmetric_contraction} already shares $\Psi$ and $\Phi$
across variables, so they now dominate the budget: $R_m(L+H)$ against only
$R_mC$ for the response. Both factors are indexed by absolute time, yet the
signal is periodic, so a position matters mainly through the cycle it falls
in and its phase within that cycle. Writing $t-1=(k-1)P+p$ for cycle index
$k$ and phase $p$, we give each temporal factor a rank-$R_k$
\textbf{C}ycl\textbf{e}-\textbf{S}eparable \textbf{T}emporal (CeST) form,
\begin{align}
\Psi_{t,m}=\sum_{r=1}^{R_k}A^{\Psi}_{k,r,m}B^{\Psi}_{p,r,m},
\label{eq:cest}
\end{align}
a Kronecker product of a cycle-index factor $A^{\Psi}\in\mathbb{R}^{K\times
R_k\times R_m}$ and a within-cycle phase factor $B^{\Psi}\in\mathbb{R}^{P
\times R_k\times R_m}$, with $K=\lceil L/P\rceil$. The cost falls from $LR_m$
to $R_kR_m(K+P)$, and the two factors separate \emph{how far back} evidence
lies from \emph{where in the cycle} it sits.

\subsubsection{History-End CeST.}
Applying Eq.~\eqref{eq:cest} to $\Psi$ gives the mode coefficients
$Z_{b,m,c}$ of Eq.~\eqref{eq:asymmetric_contraction} without materializing
$\Psi$. A strictly periodic basis cannot express a transient in the last few
observations, so we add a short dense correction over the final $M$ steps,
\begin{align}
Z_{b,m,c}=\sum_{t=1}^{L}\Psi_{t,m}\mathcal{R}^{\mathrm{hist}}_{b,t,c}
+\sum_{j=1}^{M}E_{j,m}\mathcal{R}^{\mathrm{hist}}_{b,L-M+j,c},
\label{eq:recent_window}
\end{align}
with $E\in\mathbb{R}^{M\times R_m}$ initialized to zero, so the model starts
from the purely periodic basis and departs from it only if the data demand
it. History-side cost is $R_kR_m(K+P)+MR_m$.

\subsubsection{Future-End CeST.}
The same factorization applies to $\Phi$ with $K'=\lceil H/P\rceil$ output
cycles. It is only used when the horizon spans at least one full cycle: if
$H\le P$ there is no cycle index to separate, and $\Phi$ stays dense. This
condition, rather than any property of the data, is what makes the
future-side term inactive on the short-horizon PEMS settings.

\subsubsection{Cycle-Shared Variable Response.}
The response $\mathbf{G}\in\mathbb{R}^{C\times R_m}$ is the last term that
still grows with $C$. The prototype already assigns every variable a
coordinate $\mathbf{V}_{c,:}\in\mathbb{R}^{R_p}$, so we reuse it,
\begin{align}
\mathbf{G}=\mathbf{1}+\gamma\tanh(\mathbf{V}\mathbf{M}_G),
\qquad \mathbf{M}_G\in\mathbb{R}^{R_p\times R_m},
\label{eq:cycle_shared_response}
\end{align}
where $\gamma$ bounds the modulation. This replaces $CR_m$ parameters by
$R_pR_m$. Since $R_p\le C$ by construction, the factorization is never more
expensive, and each additional variable then costs only the $R_p$ entries of
its prototype coordinate. The phase coordinates $\mathbf{U}$ supply an
analogous modulation of $\Phi$ across future phases.

\subsubsection{Residual Echo.}
The observation at the same phase one cycle earlier is a direct
local cue that the mode basis need not carry. For future step $h$ we read
$s(h)=L-P+((h-1)\bmod P)+1$ and add a phase-conditioned, cycle-decayed copy of $\mathcal{R}^{\mathrm{hist}}_{b,s(h),c}$, gated to zero whenever $s(h)<1$. The term is therefore identically zero when the window is shorter than one cycle.

\section{Experimental Results}
\label{sec:results}
\subsubsection{Datasets}
We evaluate AsyTO on datasets including: the ETT family ~\cite{zhou2021informer}, Electricity and
Traffic ~\cite{lai2018modeling}, Weather ~\cite{wu2021autoformer}, and 4
PEMS datasets~\cite{liu2022scinet}. The appendix summarizes the
statistics, including the period $P$ each dataset is assumed to have. For
the 7 non-PEMS datasets, we use a context length of $L=720$ and
forecasting horizons of $H\in\{96,192,336,720\}$. ETT follows the standard
calendar split, while Electricity, Traffic, and Weather use chronological
70\%/10\%/20\% splits. Following the standard
short-horizon traffic forecasting protocol, we evaluate the PEMS datasets
using $L=96$, $H\in\{12,24,48,96\}$, and chronological splits of
60\%/20\%/20\%.

\subsubsection{Baselines and Metrics}
We compare against 12 published forecasters, including \emph{lightweight} group: SparseTSF~\cite{lin2024sparsetsf},
MixLinear~\cite{mamixlinear}, PhaseFormer~\cite{niu2025phaseformer},
FITS~\cite{xu2024fits}, CycleNet~\cite{lin2024cyclenet},
DLinear~\cite{zeng2023transformers} and FreqCycle~\cite{zhang2026freqcycle}.
The \emph{conventional} group: FilterNet~\cite{yi2024filternet},
PatchTST~\cite{nie2022time}, TQNet~\cite{lin2025temporal},
iTransformer~\cite{liu2024itransformer} and FreTS~\cite{yi2023frequency}.
Mean squared error (MSE) and mean absolute error (MAE) on the standardized series are reported for evaluation, averaged over seeds $\{2024, 2025, 2026\}$. 
\subsubsection{Implementation Details}
All models in this work are implemented with PyTorch~\cite{paszke2019pytorch},  training uses the Adam optimizer with MSE loss. All the experiments are conducted on NVIDIA H100 GPUs.

\subsection{Main Results}
\label{sec:exp_main}
\definecolor{BestText}{HTML}{C00000}
\definecolor{SecondText}{HTML}{0000FF}

\begin{table*}[!htb]
  \centering
  \caption{The main forecasting results, averaged over 3 seeds. Left: lightweight models; right: conventional models. The best and second-best results within each model group are shown in bold red and underlined blue, respectively.}
  \label{tab:main_results}
  \setlength{\tabcolsep}{3.15pt}
  \renewcommand{\arraystretch}{1.04}
  \resizebox{0.99\textwidth}{!}{%
  \begin{tabular}{l c !{\vrule width .4pt} *{8}{cc} !{\hspace{3pt}\vrule width .9pt\hspace{3pt}} *{5}{cc}}
    \toprule
    \multirow{3}{*}{} & \multirow{3}{*}{\textbf{$H$}} & \multicolumn{16}{c!{\hspace{3pt}\vrule width .9pt\hspace{3pt}}}{\textbf{Lightweight Models}} & \multicolumn{10}{c}{\textbf{Conventional Models}} \\
    \cmidrule(lr){3-18}\cmidrule(lr){19-28}
     &   & \multicolumn{2}{c}{\textbf{Ours}} & \multicolumn{2}{c}{\textbf{SparseTSF}} & \multicolumn{2}{c}{\textbf{MixLinear}} & \multicolumn{2}{c}{\textbf{PhaseFormer}} & \multicolumn{2}{c}{\textbf{FITS}} & \multicolumn{2}{c}{\textbf{CycleNet}} & \multicolumn{2}{c}{\textbf{DLinear}} & \multicolumn{2}{c!{\hspace{3pt}\vrule width .9pt\hspace{3pt}}}{\textbf{FreqCycle}} & \multicolumn{2}{c}{\textbf{FilterNet}} & \multicolumn{2}{c}{\textbf{PatchTST}} & \multicolumn{2}{c}{\textbf{TQNet}} & \multicolumn{2}{c}{\textbf{iTrans.}} & \multicolumn{2}{c}{\textbf{FreTS}} \\
     &   & \scriptsize MSE & \scriptsize MAE & \scriptsize MSE & \scriptsize MAE & \scriptsize MSE & \scriptsize MAE & \scriptsize MSE & \scriptsize MAE & \scriptsize MSE & \scriptsize MAE & \scriptsize MSE & \scriptsize MAE & \scriptsize MSE & \scriptsize MAE & \scriptsize MSE & \scriptsize MAE & \scriptsize MSE & \scriptsize MAE & \scriptsize MSE & \scriptsize MAE & \scriptsize MSE & \scriptsize MAE & \scriptsize MSE & \scriptsize MAE & \scriptsize MSE & \scriptsize MAE \\
    \midrule
    \multirow{4}{*}{\rotatebox[origin=c]{90}{\textbf{ETTh1}}} & 96 & \textcolor{BestText}{\textbf{0.353}} & \textcolor{BestText}{\textbf{0.393}} & 0.374 & 0.405 & \textcolor{SecondText}{\underline{0.367}} & 0.401 & 0.374 & \textcolor{SecondText}{\underline{0.395}} & 0.379 & 0.402 & 0.379 & 0.403 & 0.395 & 0.419 & 0.368 & 0.401 & 0.407 & 0.435 & \textcolor{BestText}{\textbf{0.379}} & \textcolor{BestText}{\textbf{0.410}} & \textcolor{SecondText}{\underline{0.382}} & \textcolor{BestText}{\textbf{0.410}} & 0.395 & \textcolor{SecondText}{\underline{0.426}} & 0.452 & 0.462 \\
     & 192 & \textcolor{BestText}{\textbf{0.389}} & 0.415 & 0.404 & \textcolor{BestText}{\textbf{0.412}} & \textcolor{SecondText}{\underline{0.396}} & 0.419 & 0.407 & \textcolor{SecondText}{\underline{0.414}} & 0.413 & 0.423 & 0.418 & 0.427 & 0.439 & 0.450 & 0.404 & 0.427 & 0.443 & 0.453 & \textcolor{BestText}{\textbf{0.417}} & \textcolor{BestText}{\textbf{0.435}} & \textcolor{SecondText}{\underline{0.426}} & \textcolor{SecondText}{\underline{0.437}} & 0.429 & 0.449 & 0.500 & 0.495 \\
     & 336 & \textcolor{SecondText}{\underline{0.418}} & \textcolor{SecondText}{\underline{0.434}} & 0.432 & \textcolor{BestText}{\textbf{0.427}} & \textcolor{BestText}{\textbf{0.412}} & 0.436 & 0.439 & 0.435 & 0.433 & 0.439 & 0.449 & 0.447 & 0.467 & 0.470 & 0.428 & 0.441 & \textcolor{SecondText}{\underline{0.454}} & 0.466 & \textcolor{BestText}{\textbf{0.444}} & \textcolor{BestText}{\textbf{0.455}} & 0.457 & \textcolor{SecondText}{\underline{0.457}} & 0.493 & 0.493 & 0.527 & 0.510 \\
     & 720 & 0.449 & 0.469 & 0.427 & \textcolor{SecondText}{\underline{0.450}} & \textcolor{BestText}{\textbf{0.423}} & 0.453 & \textcolor{SecondText}{\underline{0.425}} & \textcolor{BestText}{\textbf{0.444}} & 0.431 & 0.457 & 0.475 & 0.482 & 0.512 & 0.524 & 0.484 & 0.493 & \textcolor{SecondText}{\underline{0.500}} & \textcolor{SecondText}{\underline{0.502}} & \textcolor{BestText}{\textbf{0.468}} & \textcolor{BestText}{\textbf{0.484}} & 0.503 & 0.507 & 0.674 & 0.598 & 0.606 & 0.562 \\
    \midrule
    \multirow{4}{*}{\rotatebox[origin=c]{90}{\textbf{ETTh2}}} & 96 & 0.273 & \textcolor{BestText}{\textbf{0.335}} & 0.286 & 0.344 & 0.287 & 0.345 & 0.281 & 0.343 & \textcolor{SecondText}{\underline{0.273}} & 0.338 & \textcolor{BestText}{\textbf{0.272}} & \textcolor{SecondText}{\underline{0.337}} & 0.323 & 0.382 & 0.282 & 0.345 & 0.327 & 0.380 & \textcolor{BestText}{\textbf{0.278}} & \textcolor{BestText}{\textbf{0.341}} & \textcolor{SecondText}{\underline{0.284}} & \textcolor{SecondText}{\underline{0.349}} & 0.306 & 0.362 & 0.366 & 0.408 \\
     & 192 & 0.340 & 0.380 & 0.343 & 0.379 & 0.340 & \textcolor{SecondText}{\underline{0.379}} & 0.343 & 0.381 & \textcolor{BestText}{\textbf{0.332}} & \textcolor{BestText}{\textbf{0.376}} & \textcolor{SecondText}{\underline{0.333}} & 0.380 & 0.408 & 0.432 & 0.345 & 0.391 & 0.381 & 0.410 & \textcolor{BestText}{\textbf{0.345}} & \textcolor{BestText}{\textbf{0.383}} & \textcolor{SecondText}{\underline{0.359}} & \textcolor{SecondText}{\underline{0.396}} & 0.403 & 0.419 & 0.442 & 0.454 \\
     & 336 & 0.375 & 0.409 & \textcolor{SecondText}{\underline{0.358}} & \textcolor{SecondText}{\underline{0.397}} & 0.359 & 0.397 & 0.375 & 0.408 & \textcolor{BestText}{\textbf{0.355}} & \textcolor{BestText}{\textbf{0.397}} & 0.365 & 0.410 & 0.495 & 0.488 & 0.387 & 0.427 & 0.400 & 0.432 & \textcolor{BestText}{\textbf{0.370}} & \textcolor{BestText}{\textbf{0.407}} & \textcolor{SecondText}{\underline{0.395}} & \textcolor{SecondText}{\underline{0.422}} & 0.460 & 0.458 & 0.608 & 0.538 \\
     & 720 & 0.399 & 0.437 & \textcolor{SecondText}{\underline{0.381}} & \textcolor{BestText}{\textbf{0.423}} & 0.382 & \textcolor{SecondText}{\underline{0.424}} & 0.421 & 0.453 & \textcolor{BestText}{\textbf{0.378}} & 0.424 & 0.416 & 0.450 & 0.833 & 0.646 & 0.441 & 0.468 & 0.422 & \textcolor{SecondText}{\underline{0.451}} & \textcolor{BestText}{\textbf{0.403}} & \textcolor{BestText}{\textbf{0.440}} & \textcolor{SecondText}{\underline{0.420}} & 0.452 & 0.433 & 0.463 & 1.202 & 0.765 \\
    \midrule
    \multirow{4}{*}{\rotatebox[origin=c]{90}{\textbf{ETTm1}}} & 96 & \textcolor{BestText}{\textbf{0.287}} & \textcolor{BestText}{\textbf{0.340}} & 0.343 & 0.374 & 0.333 & 0.374 & 0.298 & \textcolor{SecondText}{\underline{0.346}} & 0.319 & 0.360 & 0.319 & 0.362 & 0.315 & 0.359 & \textcolor{SecondText}{\underline{0.296}} & 0.351 & 0.323 & 0.372 & \textcolor{BestText}{\textbf{0.297}} & \textcolor{BestText}{\textbf{0.352}} & \textcolor{SecondText}{\underline{0.298}} & \textcolor{BestText}{\textbf{0.352}} & 0.319 & \textcolor{SecondText}{\underline{0.371}} & 0.356 & 0.392 \\
     & 192 & \textcolor{BestText}{\textbf{0.324}} & \textcolor{BestText}{\textbf{0.364}} & 0.355 & 0.379 & 0.353 & 0.383 & \textcolor{SecondText}{\underline{0.331}} & \textcolor{SecondText}{\underline{0.365}} & 0.345 & 0.374 & 0.339 & 0.373 & 0.347 & 0.381 & 0.334 & 0.372 & 0.356 & 0.389 & \textcolor{BestText}{\textbf{0.337}} & \textcolor{BestText}{\textbf{0.376}} & \textcolor{SecondText}{\underline{0.339}} & \textcolor{SecondText}{\underline{0.377}} & 0.349 & 0.389 & 0.388 & 0.410 \\
     & 336 & \textcolor{BestText}{\textbf{0.353}} & \textcolor{SecondText}{\underline{0.383}} & 0.381 & 0.393 & 0.391 & 0.408 & \textcolor{SecondText}{\underline{0.360}} & \textcolor{BestText}{\textbf{0.382}} & 0.372 & 0.390 & 0.363 & 0.386 & 0.373 & 0.394 & 0.364 & 0.389 & 0.383 & 0.400 & \textcolor{BestText}{\textbf{0.366}} & \textcolor{BestText}{\textbf{0.396}} & \textcolor{SecondText}{\underline{0.373}} & \textcolor{SecondText}{\underline{0.397}} & 0.382 & 0.410 & 0.416 & 0.427 \\
     & 720 & \textcolor{BestText}{\textbf{0.405}} & \textcolor{BestText}{\textbf{0.411}} & 0.425 & 0.416 & 0.432 & 0.429 & 0.414 & \textcolor{SecondText}{\underline{0.412}} & 0.420 & 0.416 & \textcolor{SecondText}{\underline{0.412}} & 0.413 & 0.424 & 0.425 & 0.416 & 0.415 & \textcolor{SecondText}{\underline{0.432}} & \textcolor{SecondText}{\underline{0.424}} & \textcolor{BestText}{\textbf{0.422}} & \textcolor{BestText}{\textbf{0.422}} & 0.439 & 0.431 & 0.448 & 0.448 & 0.468 & 0.460 \\
    \midrule
    \multirow{4}{*}{\rotatebox[origin=c]{90}{\textbf{ETTm2}}} & 96 & \textcolor{SecondText}{\underline{0.161}} & \textcolor{SecondText}{\underline{0.252}} & 0.176 & 0.264 & 0.167 & 0.257 & 0.174 & 0.264 & 0.167 & 0.258 & \textcolor{BestText}{\textbf{0.160}} & \textcolor{BestText}{\textbf{0.250}} & 0.178 & 0.273 & 0.168 & 0.260 & 0.184 & 0.276 & \textcolor{BestText}{\textbf{0.165}} & \textcolor{BestText}{\textbf{0.256}} & \textcolor{SecondText}{\underline{0.174}} & \textcolor{SecondText}{\underline{0.262}} & 0.182 & 0.276 & 0.180 & 0.268 \\
     & 192 & \textcolor{BestText}{\textbf{0.214}} & \textcolor{SecondText}{\underline{0.290}} & 0.226 & 0.298 & 0.221 & 0.295 & 0.227 & 0.299 & 0.221 & 0.295 & \textcolor{SecondText}{\underline{0.215}} & \textcolor{BestText}{\textbf{0.289}} & 0.282 & 0.359 & 0.222 & 0.297 & 0.238 & 0.309 & \textcolor{BestText}{\textbf{0.217}} & \textcolor{BestText}{\textbf{0.294}} & \textcolor{SecondText}{\underline{0.225}} & \textcolor{SecondText}{\underline{0.300}} & 0.249 & 0.320 & 0.248 & 0.317 \\
     & 336 & \textcolor{SecondText}{\underline{0.269}} & \textcolor{SecondText}{\underline{0.328}} & 0.277 & 0.331 & 0.271 & 0.328 & 0.274 & 0.329 & 0.272 & 0.329 & \textcolor{BestText}{\textbf{0.267}} & \textcolor{BestText}{\textbf{0.326}} & 0.296 & 0.359 & 0.275 & 0.333 & 0.286 & \textcolor{SecondText}{\underline{0.341}} & \textcolor{BestText}{\textbf{0.270}} & \textcolor{BestText}{\textbf{0.329}} & \textcolor{SecondText}{\underline{0.285}} & 0.343 & 0.303 & 0.355 & 0.308 & 0.354 \\
     & 720 & 0.353 & 0.383 & 0.356 & 0.381 & 0.353 & 0.381 & 0.353 & \textcolor{BestText}{\textbf{0.379}} & \textcolor{SecondText}{\underline{0.351}} & \textcolor{SecondText}{\underline{0.380}} & 0.354 & 0.384 & 0.414 & 0.433 & \textcolor{BestText}{\textbf{0.350}} & 0.384 & \textcolor{SecondText}{\underline{0.370}} & 0.399 & \textcolor{BestText}{\textbf{0.352}} & \textcolor{BestText}{\textbf{0.381}} & 0.373 & \textcolor{SecondText}{\underline{0.395}} & 0.376 & 0.402 & 0.380 & 0.409 \\
    \midrule
    \multirow{4}{*}{\rotatebox[origin=c]{90}{\textbf{Weather}}} & 96 & \textcolor{BestText}{\textbf{0.145}} & \textcolor{SecondText}{\underline{0.198}} & 0.180 & 0.236 & 0.175 & 0.231 & 0.148 & \textcolor{BestText}{\textbf{0.193}} & 0.174 & 0.229 & 0.165 & 0.221 & 0.172 & 0.234 & \textcolor{SecondText}{\underline{0.148}} & 0.203 & \textcolor{SecondText}{\underline{0.157}} & 0.213 & \textcolor{BestText}{\textbf{0.148}} & \textcolor{BestText}{\textbf{0.199}} & \textcolor{SecondText}{\underline{0.157}} & \textcolor{SecondText}{\underline{0.210}} & 0.178 & 0.229 & 0.159 & 0.223 \\
     & 192 & \textcolor{BestText}{\textbf{0.193}} & \textcolor{SecondText}{\underline{0.243}} & 0.221 & 0.269 & 0.218 & 0.268 & 0.194 & \textcolor{BestText}{\textbf{0.238}} & 0.215 & 0.263 & 0.211 & 0.259 & 0.215 & 0.272 & \textcolor{SecondText}{\underline{0.193}} & 0.245 & 0.207 & \textcolor{SecondText}{\underline{0.257}} & \textcolor{BestText}{\textbf{0.192}} & \textcolor{BestText}{\textbf{0.242}} & 0.208 & \textcolor{SecondText}{\underline{0.257}} & 0.226 & 0.268 & \textcolor{SecondText}{\underline{0.201}} & 0.263 \\
     & 336 & \textcolor{BestText}{\textbf{0.241}} & \textcolor{SecondText}{\underline{0.281}} & 0.263 & 0.300 & 0.264 & 0.303 & \textcolor{SecondText}{\underline{0.245}} & \textcolor{BestText}{\textbf{0.280}} & 0.260 & 0.296 & 0.255 & 0.292 & 0.258 & 0.305 & 0.247 & 0.287 & 0.262 & 0.297 & \textcolor{BestText}{\textbf{0.246}} & \textcolor{BestText}{\textbf{0.285}} & 0.258 & \textcolor{SecondText}{\underline{0.295}} & 0.291 & 0.313 & \textcolor{SecondText}{\underline{0.251}} & 0.302 \\
     & 720 & \textcolor{BestText}{\textbf{0.311}} & \textcolor{SecondText}{\underline{0.332}} & 0.324 & 0.343 & 0.327 & 0.347 & \textcolor{SecondText}{\underline{0.316}} & \textcolor{BestText}{\textbf{0.331}} & 0.321 & 0.340 & 0.321 & 0.339 & 0.319 & 0.357 & 0.321 & 0.338 & \textcolor{SecondText}{\underline{0.314}} & \textcolor{SecondText}{\underline{0.334}} & \textcolor{BestText}{\textbf{0.311}} & \textcolor{BestText}{\textbf{0.332}} & 0.335 & 0.348 & 0.358 & 0.362 & 0.315 & 0.347 \\
    \midrule
    \multirow{4}{*}{\rotatebox[origin=c]{90}{\textbf{ECL}}} & 96 & \textcolor{BestText}{\textbf{0.127}} & \textcolor{SecondText}{\underline{0.222}} & 0.139 & 0.233 & 0.150 & 0.245 & 0.129 & \textcolor{BestText}{\textbf{0.221}} & 0.142 & 0.243 & \textcolor{SecondText}{\underline{0.128}} & 0.223 & 0.133 & 0.230 & 0.129 & 0.225 & 0.142 & 0.240 & \textcolor{BestText}{\textbf{0.130}} & \textcolor{BestText}{\textbf{0.224}} & 0.136 & 0.235 & 0.134 & \textcolor{SecondText}{\underline{0.229}} & \textcolor{SecondText}{\underline{0.131}} & \textcolor{SecondText}{\underline{0.229}} \\
     & 192 & \textcolor{BestText}{\textbf{0.143}} & \textcolor{SecondText}{\underline{0.237}} & 0.151 & 0.245 & 0.163 & 0.257 & 0.146 & \textcolor{BestText}{\textbf{0.236}} & 0.156 & 0.256 & \textcolor{SecondText}{\underline{0.143}} & 0.237 & 0.147 & 0.244 & 0.148 & 0.242 & 0.170 & 0.265 & \textcolor{BestText}{\textbf{0.147}} & \textcolor{BestText}{\textbf{0.241}} & \textcolor{SecondText}{\underline{0.152}} & 0.247 & 0.153 & 0.247 & \textcolor{BestText}{\textbf{0.147}} & \textcolor{SecondText}{\underline{0.243}} \\
     & 336 & \textcolor{BestText}{\textbf{0.159}} & \textcolor{BestText}{\textbf{0.254}} & 0.166 & 0.260 & 0.178 & 0.273 & 0.167 & 0.258 & 0.172 & 0.271 & \textcolor{SecondText}{\underline{0.159}} & \textcolor{SecondText}{\underline{0.254}} & 0.162 & 0.261 & 0.162 & 0.259 & 0.187 & 0.283 & \textcolor{BestText}{\textbf{0.162}} & \textcolor{BestText}{\textbf{0.257}} & 0.171 & 0.267 & 0.167 & \textcolor{SecondText}{\underline{0.263}} & \textcolor{SecondText}{\underline{0.164}} & \textcolor{SecondText}{\underline{0.263}} \\
     & 720 & \textcolor{BestText}{\textbf{0.196}} & \textcolor{SecondText}{\underline{0.287}} & 0.205 & 0.294 & 0.217 & 0.305 & 0.200 & \textcolor{BestText}{\textbf{0.286}} & 0.210 & 0.303 & 0.197 & 0.287 & \textcolor{SecondText}{\underline{0.197}} & 0.294 & 0.198 & 0.292 & 0.221 & 0.311 & \textcolor{SecondText}{\underline{0.197}} & \textcolor{SecondText}{\underline{0.288}} & 0.198 & 0.294 & \textcolor{BestText}{\textbf{0.188}} & \textcolor{BestText}{\textbf{0.284}} & 0.200 & 0.299 \\
    \midrule
    \multirow{4}{*}{\rotatebox[origin=c]{90}{\textbf{Traffic}}} & 96 & \textcolor{SecondText}{\underline{0.379}} & \textcolor{SecondText}{\underline{0.263}} & 0.389 & 0.266 & 0.406 & 0.281 & \textcolor{BestText}{\textbf{0.363}} & \textcolor{BestText}{\textbf{0.233}} & 0.393 & 0.279 & 0.381 & 0.265 & 0.422 & 0.327 & 0.393 & 0.272 & \textcolor{SecondText}{\underline{0.355}} & 0.260 & 0.369 & \textcolor{SecondText}{\underline{0.255}} & 0.397 & 0.296 & \textcolor{BestText}{\textbf{0.345}} & \textcolor{BestText}{\textbf{0.250}} & 0.378 & 0.269 \\
     & 192 & 0.395 & \textcolor{SecondText}{\underline{0.270}} & 0.399 & 0.270 & 0.419 & 0.286 & \textcolor{BestText}{\textbf{0.378}} & \textcolor{BestText}{\textbf{0.242}} & 0.404 & 0.283 & \textcolor{SecondText}{\underline{0.394}} & 0.272 & 0.434 & 0.332 & 0.410 & 0.275 & \textcolor{SecondText}{\underline{0.371}} & 0.268 & 0.383 & \textcolor{SecondText}{\underline{0.261}} & 0.395 & 0.274 & \textcolor{BestText}{\textbf{0.360}} & \textcolor{BestText}{\textbf{0.259}} & 0.398 & 0.279 \\
     & 336 & 0.409 & 0.277 & 0.411 & \textcolor{SecondText}{\underline{0.276}} & 0.429 & 0.292 & \textcolor{BestText}{\textbf{0.397}} & \textcolor{BestText}{\textbf{0.251}} & 0.417 & 0.288 & \textcolor{SecondText}{\underline{0.406}} & 0.279 & 0.449 & 0.338 & 0.421 & 0.281 & \textcolor{SecondText}{\underline{0.386}} & 0.275 & 0.397 & \textcolor{SecondText}{\underline{0.268}} & 0.418 & 0.295 & \textcolor{BestText}{\textbf{0.369}} & \textcolor{BestText}{\textbf{0.266}} & 0.417 & 0.287 \\
     & 720 & \textcolor{SecondText}{\underline{0.440}} & 0.296 & 0.448 & \textcolor{SecondText}{\underline{0.296}} & 0.469 & 0.313 & \textcolor{BestText}{\textbf{0.434}} & \textcolor{BestText}{\textbf{0.271}} & 0.454 & 0.308 & 0.442 & 0.301 & 0.491 & 0.359 & 0.458 & 0.304 & 0.437 & 0.298 & \textcolor{SecondText}{\underline{0.434}} & \textcolor{SecondText}{\underline{0.289}} & 0.440 & 0.297 & \textcolor{BestText}{\textbf{0.394}} & \textcolor{BestText}{\textbf{0.280}} & 0.470 & 0.312 \\
    \midrule
    \multirow{4}{*}{\rotatebox[origin=c]{90}{\textbf{PEMS03}}} & 12 & \textcolor{BestText}{\textbf{0.062}} & \textcolor{BestText}{\textbf{0.167}} & 0.185 & 0.296 & 0.197 & 0.306 & 0.095 & 0.205 & 0.116 & 0.226 & 0.079 & 0.191 & 0.103 & 0.218 & \textcolor{SecondText}{\underline{0.069}} & \textcolor{SecondText}{\underline{0.175}} & 0.068 & 0.173 & 0.080 & 0.190 & \textcolor{BestText}{\textbf{0.061}} & \textcolor{BestText}{\textbf{0.161}} & \textcolor{SecondText}{\underline{0.067}} & \textcolor{SecondText}{\underline{0.171}} & 0.080 & 0.190 \\
     & 24 & \textcolor{BestText}{\textbf{0.081}} & \textcolor{BestText}{\textbf{0.188}} & 0.324 & 0.395 & 0.429 & 0.474 & 0.156 & 0.265 & 0.234 & 0.323 & 0.121 & 0.238 & 0.180 & 0.293 & \textcolor{SecondText}{\underline{0.100}} & \textcolor{SecondText}{\underline{0.212}} & 0.096 & 0.204 & 0.131 & 0.244 & \textcolor{BestText}{\textbf{0.078}} & \textcolor{BestText}{\textbf{0.183}} & \textcolor{SecondText}{\underline{0.093}} & \textcolor{SecondText}{\underline{0.202}} & 0.123 & 0.236 \\
     & 48 & \textcolor{BestText}{\textbf{0.115}} & \textcolor{BestText}{\textbf{0.219}} & 0.650 & 0.586 & 0.776 & 0.661 & 0.292 & 0.368 & 0.542 & 0.522 & \textcolor{SecondText}{\underline{0.156}} & \textcolor{SecondText}{\underline{0.257}} & 0.317 & 0.407 & 0.167 & 0.276 & \textcolor{SecondText}{\underline{0.149}} & \textcolor{SecondText}{\underline{0.258}} & 0.232 & 0.330 & \textcolor{BestText}{\textbf{0.107}} & \textcolor{BestText}{\textbf{0.215}} & 0.151 & 0.260 & 0.196 & 0.305 \\
     & 96 & \textcolor{BestText}{\textbf{0.156}} & \textcolor{BestText}{\textbf{0.249}} & 1.186 & 0.846 & 1.352 & 0.927 & 0.513 & 0.502 & 1.063 & 0.791 & \textcolor{SecondText}{\underline{0.199}} & \textcolor{SecondText}{\underline{0.292}} & 0.451 & 0.507 & 0.256 & 0.351 & \textcolor{SecondText}{\underline{0.228}} & \textcolor{SecondText}{\underline{0.326}} & 0.385 & 0.444 & \textcolor{BestText}{\textbf{0.149}} & \textcolor{BestText}{\textbf{0.254}} & 0.319 & 0.392 & 0.262 & 0.362 \\
    \midrule
    \multirow{4}{*}{\rotatebox[origin=c]{90}{\textbf{PEMS04}}} & 12 & \textcolor{BestText}{\textbf{0.075}} & \textcolor{BestText}{\textbf{0.180}} & 0.197 & 0.312 & 0.209 & 0.322 & 0.113 & 0.226 & 0.130 & 0.241 & 0.089 & 0.200 & 0.114 & 0.228 & \textcolor{SecondText}{\underline{0.081}} & \textcolor{SecondText}{\underline{0.188}} & \textcolor{SecondText}{\underline{0.078}} & \textcolor{SecondText}{\underline{0.183}} & 0.102 & 0.215 & \textcolor{BestText}{\textbf{0.066}} & \textcolor{BestText}{\textbf{0.165}} & 0.085 & 0.188 & 0.096 & 0.207 \\
     & 24 & \textcolor{BestText}{\textbf{0.090}} & \textcolor{BestText}{\textbf{0.200}} & 0.336 & 0.412 & 0.440 & 0.492 & 0.195 & 0.301 & 0.249 & 0.341 & 0.127 & 0.244 & 0.188 & 0.300 & \textcolor{SecondText}{\underline{0.113}} & \textcolor{SecondText}{\underline{0.225}} & \textcolor{SecondText}{\underline{0.097}} & \textcolor{SecondText}{\underline{0.209}} & 0.159 & 0.272 & \textcolor{BestText}{\textbf{0.077}} & \textcolor{BestText}{\textbf{0.180}} & 0.116 & 0.223 & 0.142 & 0.257 \\
     & 48 & \textcolor{BestText}{\textbf{0.115}} & \textcolor{BestText}{\textbf{0.227}} & 0.675 & 0.608 & 0.808 & 0.684 & 0.371 & 0.423 & 0.570 & 0.542 & 0.187 & 0.305 & 0.319 & 0.405 & \textcolor{SecondText}{\underline{0.179}} & \textcolor{SecondText}{\underline{0.288}} & \textcolor{SecondText}{\underline{0.135}} & \textcolor{SecondText}{\underline{0.253}} & 0.278 & 0.370 & \textcolor{BestText}{\textbf{0.096}} & \textcolor{BestText}{\textbf{0.204}} & 0.177 & 0.280 & 0.224 & 0.331 \\
     & 96 & \textcolor{BestText}{\textbf{0.143}} & \textcolor{BestText}{\textbf{0.252}} & 1.269 & 0.887 & 1.429 & 0.959 & 0.659 & 0.585 & 1.166 & 0.838 & \textcolor{SecondText}{\underline{0.190}} & \textcolor{SecondText}{\underline{0.294}} & 0.424 & 0.481 & 0.270 & 0.367 & \textcolor{SecondText}{\underline{0.201}} & \textcolor{SecondText}{\underline{0.316}} & 0.446 & 0.489 & \textcolor{BestText}{\textbf{0.124}} & \textcolor{BestText}{\textbf{0.232}} & 0.271 & 0.357 & 0.288 & 0.382 \\
    \midrule
    \multirow{4}{*}{\rotatebox[origin=c]{90}{\textbf{PEMS07}}} & 12 & \textcolor{BestText}{\textbf{0.058}} & \textcolor{BestText}{\textbf{0.155}} & 0.179 & 0.300 & 0.190 & 0.308 & 0.091 & 0.203 & 0.109 & 0.221 & 0.073 & 0.180 & 0.100 & 0.214 & \textcolor{SecondText}{\underline{0.064}} & \textcolor{SecondText}{\underline{0.165}} & \textcolor{SecondText}{\underline{0.061}} & \textcolor{SecondText}{\underline{0.160}} & 0.075 & 0.183 & \textcolor{BestText}{\textbf{0.056}} & \textcolor{BestText}{\textbf{0.153}} & 0.064 & 0.161 & 0.077 & 0.184 \\
     & 24 & \textcolor{BestText}{\textbf{0.077}} & \textcolor{BestText}{\textbf{0.177}} & 0.328 & 0.407 & 0.437 & 0.489 & 0.160 & 0.270 & 0.229 & 0.326 & 0.102 & 0.207 & 0.192 & 0.304 & \textcolor{SecondText}{\underline{0.094}} & \textcolor{SecondText}{\underline{0.202}} & \textcolor{SecondText}{\underline{0.085}} & \textcolor{SecondText}{\underline{0.188}} & 0.120 & 0.231 & \textcolor{BestText}{\textbf{0.064}} & \textcolor{BestText}{\textbf{0.160}} & 0.090 & 0.192 & 0.126 & 0.238 \\
     & 48 & \textcolor{BestText}{\textbf{0.119}} & \textcolor{BestText}{\textbf{0.208}} & 0.675 & 0.606 & 0.806 & 0.682 & 0.329 & 0.393 & 0.546 & 0.529 & 0.154 & \textcolor{SecondText}{\underline{0.253}} & 0.381 & 0.440 & \textcolor{SecondText}{\underline{0.152}} & 0.263 & \textcolor{SecondText}{\underline{0.125}} & \textcolor{SecondText}{\underline{0.232}} & 0.212 & 0.310 & \textcolor{BestText}{\textbf{0.084}} & \textcolor{BestText}{\textbf{0.184}} & 0.137 & 0.239 & 0.224 & 0.320 \\
     & 96 & \textcolor{BestText}{\textbf{0.182}} & \textcolor{BestText}{\textbf{0.238}} & 1.252 & 0.881 & 1.388 & 0.943 & 0.607 & 0.552 & 1.094 & 0.797 & \textcolor{SecondText}{\underline{0.204}} & \textcolor{SecondText}{\underline{0.291}} & 0.580 & 0.542 & 0.248 & 0.343 & \textcolor{SecondText}{\underline{0.180}} & \textcolor{SecondText}{\underline{0.287}} & 0.344 & 0.403 & \textcolor{BestText}{\textbf{0.106}} & \textcolor{BestText}{\textbf{0.206}} & 0.279 & 0.362 & 0.323 & 0.392 \\
    \midrule
    \multirow{4}{*}{\rotatebox[origin=c]{90}{\textbf{PEMS08}}} & 12 & \textcolor{BestText}{\textbf{0.080}} & \textcolor{BestText}{\textbf{0.183}} & 0.188 & 0.304 & 0.201 & 0.313 & 0.105 & 0.214 & 0.127 & 0.241 & 0.091 & 0.201 & 0.113 & 0.226 & \textcolor{SecondText}{\underline{0.081}} & \textcolor{SecondText}{\underline{0.187}} & \textcolor{SecondText}{\underline{0.078}} & 0.179 & 0.099 & 0.207 & \textcolor{BestText}{\textbf{0.070}} & \textcolor{BestText}{\textbf{0.169}} & \textcolor{SecondText}{\underline{0.078}} & \textcolor{SecondText}{\underline{0.178}} & 0.095 & 0.202 \\
     & 24 & \textcolor{BestText}{\textbf{0.117}} & \textcolor{BestText}{\textbf{0.222}} & 0.323 & 0.404 & 0.426 & 0.485 & 0.180 & 0.282 & 0.252 & 0.348 & 0.143 & 0.253 & 0.197 & 0.301 & \textcolor{SecondText}{\underline{0.119}} & \textcolor{SecondText}{\underline{0.227}} & \textcolor{SecondText}{\underline{0.108}} & 0.213 & 0.167 & 0.272 & \textcolor{BestText}{\textbf{0.094}} & \textcolor{BestText}{\textbf{0.195}} & 0.111 & \textcolor{SecondText}{\underline{0.211}} & 0.150 & 0.255 \\
     & 48 & \textcolor{BestText}{\textbf{0.190}} & \textcolor{BestText}{\textbf{0.286}} & 0.678 & 0.609 & 0.817 & 0.689 & 0.353 & 0.404 & 0.604 & 0.568 & 0.264 & 0.340 & 0.389 & 0.431 & \textcolor{SecondText}{\underline{0.202}} & \textcolor{SecondText}{\underline{0.296}} & \textcolor{SecondText}{\underline{0.170}} & \textcolor{SecondText}{\underline{0.264}} & 0.314 & 0.384 & \textcolor{BestText}{\textbf{0.152}} & \textcolor{BestText}{\textbf{0.247}} & 0.179 & 0.266 & 0.247 & 0.334 \\
     & 96 & \textcolor{SecondText}{\underline{0.336}} & \textcolor{SecondText}{\underline{0.379}} & 1.346 & 0.900 & 1.509 & 0.974 & 0.703 & 0.569 & 1.322 & 0.893 & \textcolor{BestText}{\textbf{0.281}} & \textcolor{BestText}{\textbf{0.335}} & 0.667 & 0.542 & 0.371 & 0.394 & \textcolor{SecondText}{\underline{0.285}} & \textcolor{SecondText}{\underline{0.329}} & 0.561 & 0.521 & \textcolor{BestText}{\textbf{0.260}} & \textcolor{BestText}{\textbf{0.310}} & 0.326 & 0.361 & 0.356 & 0.399 \\
    \midrule
    \multicolumn{2}{l!{\vrule width .4pt}}{$1^{st}$Cnt$^{*}$} & \textbf{30} & \textbf{21} & 0 & 3 & 2 & 0 & 4 & 14 & 3 & 2 & 4 & 4 & 0 & 0 & 1 & 0 & -- & -- & -- & -- & -- & -- & -- & -- & -- & -- \\
    \addlinespace[1pt]
    \multicolumn{2}{l!{\vrule width .4pt}}{$1^{st}$Cnt$^{\dagger}$} & 13 & 7 & 0 & 3 & 2 & 0 & 0 & 13 & 3 & 2 & 3 & 3 & 0 & 0 & 1 & 0 & 0 & 0 & 1 & 0 & \textbf{16} & \textbf{15} & 5 & 1 & 0 & 0 \\
    \bottomrule
  \end{tabular}%
  }
  \par\smallskip
  {\footnotesize\raggedright
  \textcolor{BestText}{\textbf{Best}} and \textcolor{SecondText}{\underline{Second-best}} within each model group. For the lightweight group, the original source annotations retain tie-breaking based on unrounded values.
  $1^{st}$Cnt$^{*}$ within the lightweight group, counted per metric over all settings, ties are broken on unrounded values.
  $1^{st}$Cnt$^{\dagger}$ among all compared models.\par}
\end{table*}

Table~\ref{tab:main_results} reports MSE and MAE across all 11 benchmarks. Within the lightweight group, AsyTO attains the lowest MSE in $30$ of the $44$ cells and the lowest MAE in $21$, and its mean rank among the eight lightweight models is $1.96$ on the 7 standard benchmarks and $1.06$ on PEMS. 
The advantage is concentrated where the periodic structure the model assumes is actually present, and it holds across horizons rather than only short ones. AsyTO is best across all four horizons on ETTm1, Weather, and Electricity, and in $15$ of the $16$ PEMS cells (where the regime with the shortest look-back relative to the period is least identifiable from the data). On ETTh2 the lightweight methods are separated by less than the seed-to-seed spread, and the leader changes with the horizon, so no method can be said to win there.

Traffic is the one benchmark where AsyTO loses at every horizon, trailing PhaseFormer by $4.5\%$, $4.5\%$, $2.9\%$ and $1.3\%$ as the horizon grows. This is the expected cost of the target-only design: Traffic has $862$ sensors whose predictive information is largely shared, and a model that forecasts each variable from its own history alone cannot recover it. The gap narrows as the horizon lengthens, consistent with cross-sensor information being most useful at short range. We treat this as a scope boundary rather than a tuning failure, and return to it in the discussion.

Against the conventional group AsyTO remains competitive without matching their budget: it attains the lowest MSE of all thirteen models in $13$ of the $44$ cells while using two to three orders of magnitude fewer parameters, which is the trade-off Figure~\ref{fig:eff_analysis} makes explicit.
\subsection{Efficiency Analysis}
\begin{figure*}
    \centering
    \includegraphics[width=\linewidth]{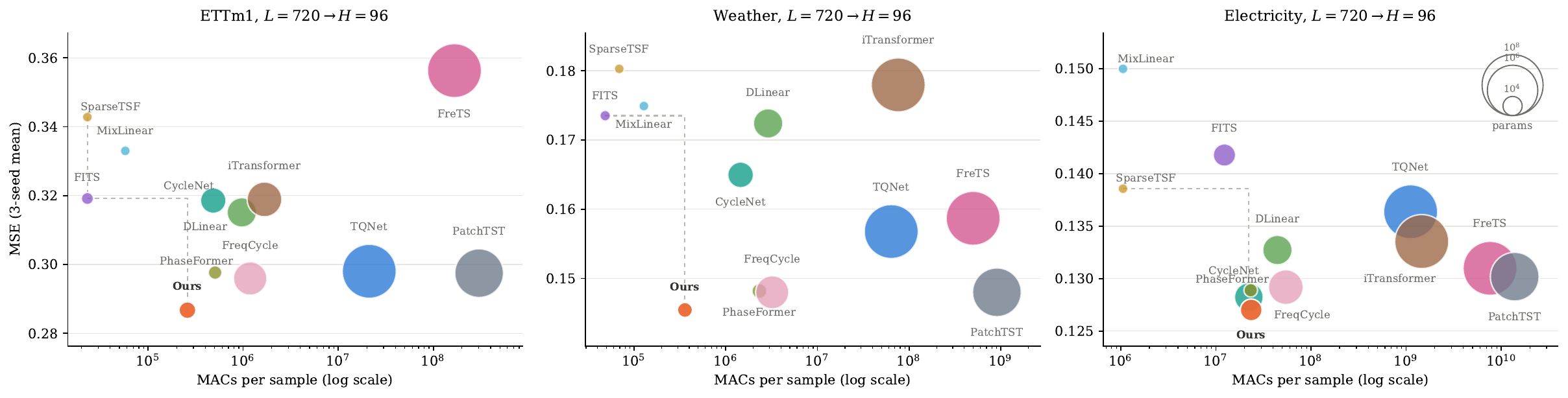}
    \caption{Visualization of accuracy vs. compute Pareto on ETTm1, Weather, Electricity under the setting of $L=720\rightarrow H=96$. The bubble size indicates the scale of the trainable parameters, where the dashed step in grey indicates the Pareto frontier.}
    \label{fig:eff_analysis}
\end{figure*}

Figure~\ref{fig:eff_analysis} places AsyTO and baselines on the
accuracy-compute plane at $L{=}720\to H{=}96$, using multiply accumulate
operations (MACs) per forward sample as a hardware-agnostic measure of
compute and trainable parameters as bubble area. 
AsyTO sits on the Pareto frontier, and it does so at a compute scale set by the cheap end of the field rather than the expensive one: on Electricity, it uses $47.5\times$, $62.2\times$, and $590.4\times$ fewer MACs than TQNet, iTransformer, and PatchTST, and on Weather the gap to PatchTST reaches $2527\times$. 
The parameter gap is wider: $92.5\times$ against TQNet and $155\times$ against iTransformer on Electricity, indicating the conventional models carry per-variable or per-token machinery that AsyTO replaces with shared factors.

The frontier is crowded at its cheap end, and AsyTO does not dominate on every axis. PhaseFormer matches our compute on Electricity ($1.0\times$ MACs) with a tenth of the parameters, and CycleNet is marginally cheaper ($0.9\times$ MACs); both trail in error by $1.5\%$ and $1.0\%$. The claim supported by the figure is therefore specific: among models at this compute scale AsyTO is the most accurate, not the smallest.

What does scale favourably is the cost of an additional variable.
Instantiating the shipped configuration at $C$ and $C{+}1$ variables shows the parameter count growing by exactly $R_p$ per variable: $2$ on Weather, $4$ on ETTm1, $8$ on Traffic and $16$ on Electricity. This is because the temporal factors are shared and the variable response is expressed in the prototype's coordinates rather than stored per variable. A dense per-variable temporal map would instead add $HL$ coefficients per variable, or $69{,}120$ at $720\!\to\!96$.

\subsection{Validating the Structural Priors}
\begin{table}[!htb]
  \centering
  \caption{Module ablation for structure prior validation.}
  \label{tab:necessity}
  \tiny
  \setlength{\tabcolsep}{5pt}
  \renewcommand{\arraystretch}{0.95}
  \resizebox{0.77\linewidth}{!}{%
\begin{tabular}{@{}lcccc@{}}
\toprule
Dataset & w/o $\mathcal{Q}$ & w/o ATO & w/o Rct. & AsyTO \\
\midrule
ETTh1 & 0.406 & 0.468 & 0.407 & \textbf{0.402} \\
ETTh2 & 0.348 & 0.397 & 0.354 & \textbf{0.346} \\
ETTm1 & 0.347 & 0.352 & 0.344 & \textbf{0.342} \\
ETTm2 & 0.251 & 0.253 & \textbf{0.249} & \textbf{0.249} \\
Weather & 0.221 & \textbf{0.218} & 0.223 & 0.223 \\
ECL & 0.159 & 0.161 & 0.157 & \textbf{0.156} \\
Traffic & \textbf{0.406} & 0.428 & 0.410 & \textbf{0.406} \\
\midrule
PEMS03 & 0.130 & 0.134 & 0.105 & \textbf{0.104} \\
PEMS04 & 0.142 & 0.133 & 0.107 & \textbf{0.106} \\
PEMS07 & 0.131 & 0.128 & \textbf{0.109} & \textbf{0.109} \\
PEMS08 & 0.219 & 0.232 & 0.182 & \textbf{0.181} \\
\bottomrule
\end{tabular}%
  }
\end{table}

\begin{figure}[t]
  \centering
  \includegraphics[width=0.8\columnwidth]{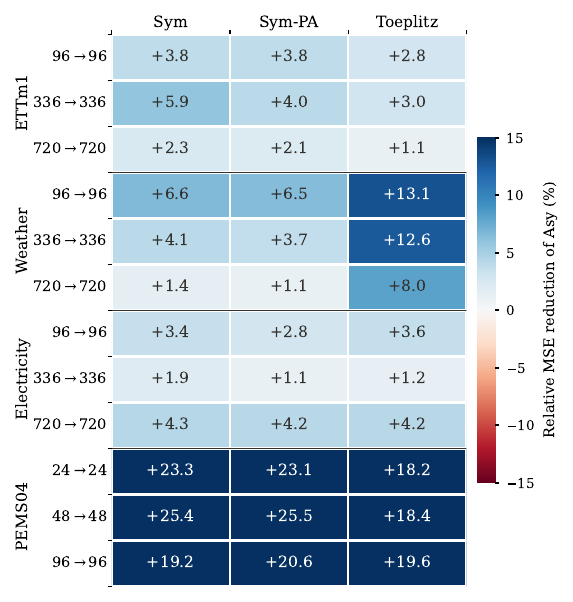}
  \caption{Directionality control. Each cell is the 3-seed averaged relative MSE reduction of \emph{Asym} ( $\boldsymbol{\Psi}\neq\boldsymbol{\Phi}$) in percentage, against 3 constrained alternatives:
  \emph{Sym} sets $\boldsymbol{\Phi}=\boldsymbol{\Psi}$ with a parameter aligned version \emph{Sym-PA}, and shift-invariant \emph{Toeplitz}. Blue means \emph{Asym} is better.}
  \label{fig:operator_asy}
\end{figure}

To validate the sources of AsyTO's performance gain, we design two complementary experiments: module ablations in Table~\ref{tab:necessity} and controlled operator constraints in Figure~\ref{fig:operator_asy}.
Table~\ref{tab:necessity} replaces each stage with a simpler alternative while retaining the training protocol. \emph{w/o $\mathcal{Q}$} applies the operator to the raw series instead of the periodicity-exclusive residual, \emph{w/o ATO} replaces the factorized operator by an independent dense $H\times L$ map per variable, \emph{w/o $G$} forces every variable to share one response, and \emph{w/o recent} removes the dense correction over the latest observations. Entries are raw test MSE, 3-seed means over 4 horizons. The three interventions reveal different roles. Periodic removal changes MSE only marginally on the seven long-context datasets, but improves it by $20$--$34\%$ on PEMS, where the look-back is shorter than the dominant cycle. The periodic prototype is therefore a regime-specific necessity rather than a universal source of improvement. In contrast, the dense replacement is worse on ten of eleven datasets despite using $26$--$19{,}509\times$ more parameters, showing that unstructured per-variable capacity cannot replace shared residual structure. Recent correction has improved the displayed averages in most cases, but its effect is at most $2.3\%$, identifying it as a secondary refinement.

The dense ablation replaces the entire residual operator and therefore cannot isolate history–future asymmetry. Figure~\ref{fig:operator_asy} provides this control using the same operator family at $L{=}H$. \emph{Asym} learns separate history and future factors. \emph{Sym} ties them by setting $\boldsymbol{\Phi}=\boldsymbol{\Psi}$, testing the common-basis constraint but using fewer parameters. \emph{Sym-PA} reallocates the saved capacity to match the parameter count of \emph{Asym}, separating directionality from model size. \emph{Toeplitz} instead makes the map depend only on relative lag, testing whether a shift-invariant rule is sufficient. Every displayed comparison favors separate factors, by $1.1$--$25.5\%$. After averaging the tested window lengths, \emph{Asym} wins $29$ of $33$ controlled setting comparisons across all datasets (see Appendix for more details), the 4 reversals occur only on ETTh1/2 and are within $1.4\%$. The persistence of the gap against \emph{Sym-PA} rules out parameter count as its explanation, while the Toeplitz comparison shows that relative lag alone cannot replace distinct history-reading and future-writing structure.
\begin{figure*}[t]
    \centering
    \includegraphics[width=.49\textwidth]{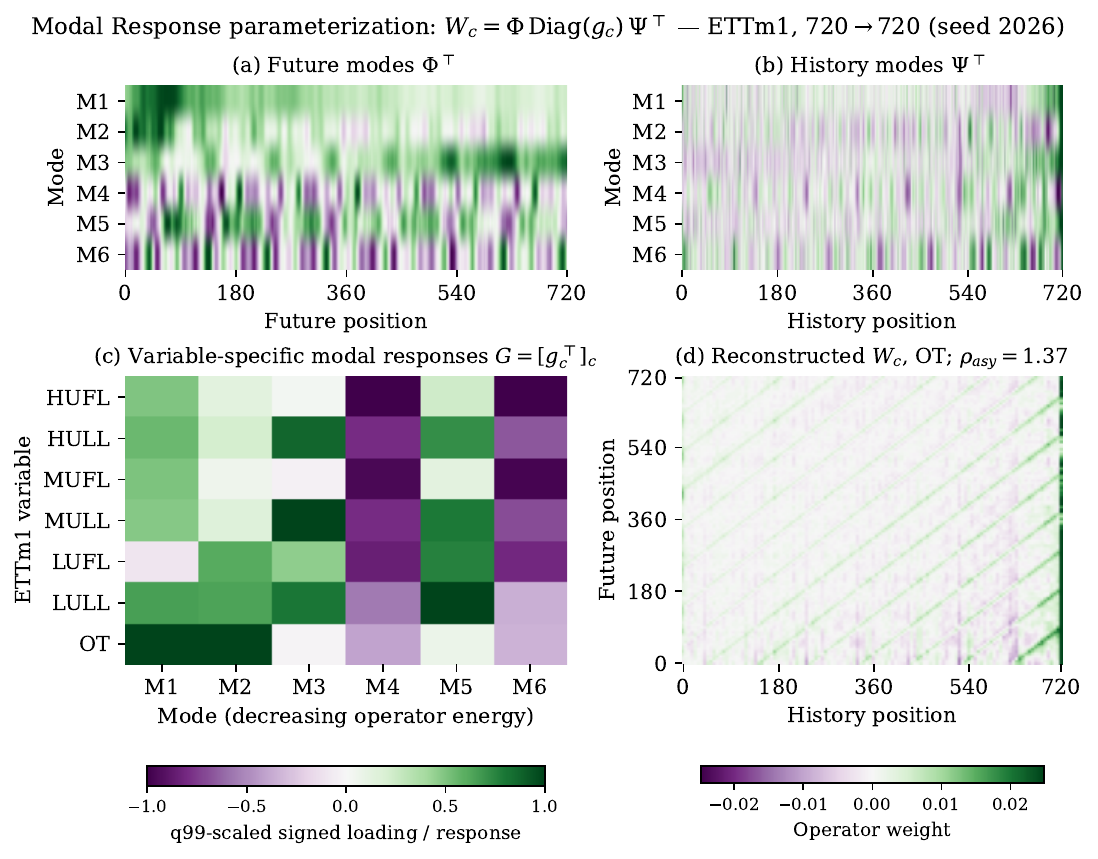}\hfill
    \includegraphics[width=.49\textwidth]{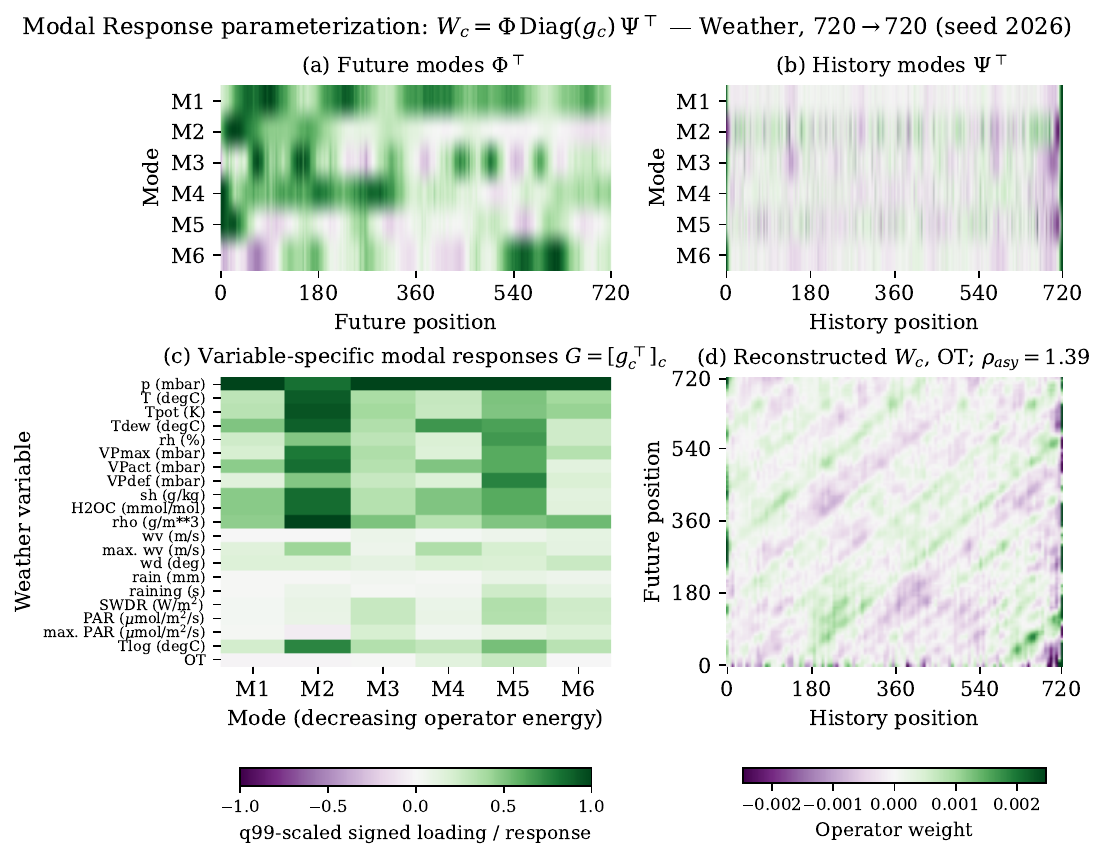}
    \caption{The learned factorization on ETTm1 and Weather at $720\!\to\!720$. (a, b) The six highest-energy future-writing and history-reading modes, (c) the variable mode responses, (d) the operator reconstructed for variables, with $\rho_{\mathrm{asy}}=\lVert\mathcal{W}-\mathcal{W}^{\top}\rVert_F/
\lVert\mathcal{W}\rVert_F$.}
    \label{fig:e4_factors}
\end{figure*}
Figure~\ref{fig:e4_factors} shows the asymmetry mechanism, where the history modes are localized in time while future modes spread across the horizon, indicating the two temporal bases are specialized rather than converge. Panel (d) puts a number on the gap: for each
dataset's target variable $\rho_{\mathrm{asy}}$ is $1.37$ on ETTm1 and $1.39$ on Weather, placing $46.9\%$ and $48.3\%$ of the operator's energy in its antisymmetric part. Panel (c) illustrates the responses grouping thermo-dynamically related variables, showing that shared modes are adapted across the panel rather than duplicated.
\subsection{Cost of Factorization}
The preceding experiments establish which structures are needed, this section answers whether their compressed parameterizations discard useful capacity.
Table~\ref{tab:factorization} changes one storage choice at a time at
$L=H=720$. \emph{Shared} variants remove variable specificity,
\emph{Independent} variants store a separate prototype or response for every variable, and \emph{Dense} variants replace one cycle-separable temporal factor with its unrestricted counterpart.
\paragraph{Variable-side Storage.}
The benefit of amortization increases with panel width. Relative to AsyTO, storing an independent prototype $Q_c$ increases the total parameter count by $1.06\times$, $1.45\times$, $2.41\times$, and $7.32\times$ on ETTm1, Weather, ECL, and Traffic, respectively, without improving the reported MSE. Independent responses $G_c$ follow the same scaling trend ($1.01$--$2.19\times$) and match or underperform AsyTO except on Weather, where they reduce MSE from $.311$ to $.307$ for a $4.5\%$ increase in parameters. Full sharing is not a uniform substitute either: it helps on Weather but degrades ETTm1 and ECL. The coordinate-generated form therefore
provides the useful middle ground between a channel-homogeneous model and costly per-variable storage, with Weather marking a possible capacity boundary when independent responses are inexpensive.
\paragraph{Temporal Storage.}
Replacing either temporal factor with a dense one raises the total model size by $1.63$--$2.90\times$, while changing MSE by at most $.003$ and with no consistent direction. Dense factors slightly improve ECL and Traffic, but match or underperform AsyTO on ETTm1 and Weather. The cycle-separable form therefore removes $39$--$65\%$ of the parameters required by its dense counterpart while retaining nearly all of its accuracy. The two factorizations address complementary scaling terms: variable-side
compression prevents per-variable storage from dominating wide panels, whereas temporal compression controls the shared history--future cost and is most visible on narrow panels. Both are therefore required for efficiency across panel widths.

\begin{table}[!h]
  \centering
  \caption{Cost of factorization at $L{=}720\rightarrow H{=}720$, the one
  horizon at which every arm exists. Each row stores one component
  differently; \emph{Par.} is its trainable-parameter count. All eleven
  datasets and horizons are in the appendix.}
  \label{tab:factorization}
  \small
  \setlength{\tabcolsep}{2pt}
  \resizebox{\linewidth}{!}{%
\begin{tabular}{@{}lcccccccc@{}}
\toprule
\multirow{2}{*}{Parameterization} & \multicolumn{2}{c}{ETTm1} & \multicolumn{2}{c}{Weather} & \multicolumn{2}{c}{Electricity} & \multicolumn{2}{c}{Traffic} \\
\cmidrule(lr){2-3} \cmidrule(lr){4-5} \cmidrule(lr){6-7} \cmidrule(lr){8-9}
 & MSE & Par. & MSE & Par. & MSE & Par. & MSE & Par. \\
\midrule
\multicolumn{9}{l}{\emph{Low-Rank Periodic Prototype} $\mathcal{Q}_s = \mathbf{U}\mathbf{V}^\top$} \\
\midrule
Shared $\mathcal{Q}$ & 0.411 & 10,498 & 0.309 & 6,846 & 0.200 & 38,346 & 0.440 & 23,074 \\
Indep. $\mathcal{Q}_c$ & 0.410 & 11,074 & 0.311 & 9,726 & 0.196 & 92,106 & 0.446 & 167,722 \\
\midrule
\multicolumn{9}{l}{\emph{Cycle-Shared Variable Response} $\mathbf{G}$} \\
\midrule
Shared $\mathbf{G}_s$ & 0.406 & 10,281 & 0.310 & 6,691 & 0.197 & 37,475 & 0.440 & 23,512 \\
Indep. $\mathbf{G}_c$ & 0.405 & 10,498 & 0.307 & 7,006 & 0.196 & 57,698 & 0.441 & 50,234 \\
\midrule
\multicolumn{9}{l}{\emph{History-CeST Factorization} $\boldsymbol{\Psi}$} \\
\midrule
Dense $\boldsymbol{\Psi}_{dense}$ & 0.407 & 27,042 & 0.312 & 14,302 & 0.195 & 67,042 & 0.439 & 37,338 \\
\midrule
\multicolumn{9}{l}{\emph{Future-CeST Factorization} $\boldsymbol{\Phi}$} \\
\midrule
Dense $\boldsymbol{\Phi}_{dense}$ & 0.407 & 30,114 & 0.311 & 15,838 & 0.195 & 73,186 & 0.437 & 40,410 \\
\midrule
AsyTO & 0.405 & 10,402 & 0.311 & 6,702 & 0.196 & 38,178 & 0.440 & 22,906 \\
\bottomrule
\end{tabular}%
  }
\end{table}

\begin{table}[!h]
  \centering
  \caption{Transferability analysis on frozen PhaseFormer and MixLinear. $T0$ is the backbone MSE, $T1_\Delta$ and $T2_\Delta$ are relative MSE changes after adding $Q$ and $Q+\mathrm{CeST}$. Negative is better, performance gains exceeding $5\%$ are bold.}
  \label{tab:transfer}
 \resizebox{\linewidth}{!}{%
\begin{tabular}{@{}l ccc ccc@{}}
\toprule
\multirow{2}{*}{Dataset} & \multicolumn{3}{c}{PhaseFormer}
 & \multicolumn{3}{c}{MixLinear} \\
\cmidrule(lr){2-4}\cmidrule(lr){5-7}
 & T0 & T1$\Delta$ & T2$\Delta$ & T0 & T1$\Delta$ & T2$\Delta$ \\
\midrule
ETTh1 & 0.419 & -0.8\% & -0.2\% & 0.400 & +3.0\% & +2.7\% \\
ETTh2 & 0.354 & -0.0\% & -0.2\% & 0.342 & +0.2\% & +0.1\% \\
ETTm1 & 0.350 & -0.2\% & -0.9\% & 0.377 & -0.4\% & \textbf{-8.1\%} \\
ETTm2 & 0.257 & -0.7\% & -2.6\% & 0.253 & -1.6\% & -1.8\% \\
Weather & 0.225 & +0.1\% & -1.7\% & 0.249 & -0.4\% & \textbf{-13.4\%} \\
ECL & 0.160 & -1.8\% & -3.0\% & 0.174 & -4.8\% & \textbf{-9.4\%} \\
Traffic & 0.393 & -0.6\% & -0.7\% & 0.417 & -1.7\% & -2.5\% \\
\midrule
PEMS03 & 0.176 & \textbf{-12.1\%} & \textbf{-24.7\%} & 0.587 & \textbf{-43.1\%} & \textbf{-70.7\%} \\
PEMS04 & 0.192 & \textbf{-16.9\%} & \textbf{-30.0\%} & 0.619 & \textbf{-30.7\%} & \textbf{-66.2\%} \\
PEMS07 & 0.208 & \textbf{-14.9\%} & \textbf{-31.8\%} & 0.609 & \textbf{-45.1\%} & \textbf{-72.6\%} \\
PEMS08 & 0.363 & \textbf{-16.7\%} & \textbf{-35.6\%} & 0.634 & \textbf{-30.9\%} & \textbf{-64.3\%} \\
\bottomrule
\end{tabular}%
 }
\end{table}

\subsection{Transferability Analysis}
Ablations establish that the proposed components are useful within AsyTO, but not whether they remain effective outside the architecture for which they were designed. We therefore attach them to two independently trained lightweight forecasters, PhaseFormer~\cite{niu2025phaseformer} and MixLinear~\cite{mamixlinear}, while keeping all backbone weights frozen. $T0$ is the frozen backbone, $T1$ trains only the periodic prototype $Q$, and $T2$ trains $Q$ together with CeST.

$T2$ reduces MSE in $39$ of $44$ PhaseFormer cells and $40$ of $44$ MixLinear cells. Across the seven standard benchmarks, the mean reductions are $1.3\%$ and $4.6\%$, respectively; on PEMS, where the look-back is shorter than one cycle ($L=96<P=288$), they increase to $30.5\%$ and $68.4\%$. This contrast is consistent with the periodic path being most useful when the backbone cannot observe a complete cycle, although the ratio $L/P$ is not the only factor that differs between these datasets.

CeST also contributes beyond periodic removal: $T2$ outperforms $T1$ on $21$ of the $22$ displayed backbone--dataset pairs. On PEMS, it raises the mean reduction from $15.2\%$ to $30.5\%$ for PhaseFormer and from $37.5\%$ to $68.4\%$ for MixLinear. The only material regression relative to $T0$ is MixLinear on ETTh1 ($+2.7\%$), where the long input already spans many cycles. Because these gains are obtained without updating any backbone weight, they demonstrate that the modules provide a transferable correction rather than relying on co-adaptation with the AsyTO architecture.
\section{Discussion and Conclusion}
This work presented AsyTO, a compact forecasting operator that assigns distinct parameterizations to history-side evidence extraction and future-side trajectory generation. After estimating and removing a low-rank periodic component, AsyTO maps the residual through separate history and future factors.
Cycle-separable factorization controls the temporal cost, while
prototype-derived coordinates generate variable-specific responses. Matched-budget controls substantiate the asymmetric design, and AsyTO achieves the lowest error among lightweight forecasters in 30 of 44 settings. Rather than using either one temporal map for all variables or an independent dense map for each, AsyTO shares temporal factors across the panel and specializes their effects through variable-specific responses. Each forecast, however, uses only the history of its target variable. This intermediate design preserves variable specificity with modest parameter growth, but cannot exploit predictive signals available exclusively from other variables. Selective cross-variable interaction is therefore a natural direction for future work.

\bibliography{ref}

\end{document}